\documentclass[conference]{IEEEtran}
\IEEEoverridecommandlockouts
\usepackage{cite}
\usepackage{amsmath,amssymb,amsfonts}
\usepackage{graphicx}
\usepackage{textcomp}
\usepackage{xcolor}
\usepackage{bm}
\usepackage{amsmath}
\usepackage{amssymb}
\usepackage{amsthm}
\usepackage{mathrsfs}
\usepackage{enumerate}
\usepackage{multirow}
\usepackage{color}
\usepackage{threeparttable}
\usepackage{subfigure}

\usepackage[square, comma, sort&compress, numbers]{natbib}
\usepackage[pagebackref=false,breaklinks=true,letterpaper=true,colorlinks,bookmarks=false]{hyperref}
\usepackage{times}
\usepackage{breakurl}
\usepackage{array}
\usepackage{verbatim}
\usepackage{algorithm}
\usepackage{bbding}
\usepackage{algpseudocode}
\usepackage{lettrine}

\usepackage{booktabs}
\usepackage{multirow}
\usepackage{colortbl} 

\definecolor{dc1}{RGB}{0,0,255} 
\definecolor{dc2}{RGB}{0,165,255}
\definecolor{dc3}{RGB}{0,255,255}
\definecolor{dc4}{RGB}{0,255,128}
\definecolor{dc5}{RGB}{0,255,0}
\definecolor{dc6}{RGB}{128,255,0}
\definecolor{dc7}{RGB}{255,255,0}
\definecolor{dc8}{RGB}{255,128,0}
\definecolor{dc9}{RGB}{255,0,0}
\definecolor{dc10}{RGB}{226,43,138}
\definecolor{dc11}{RGB}{230,216,173}
\definecolor{dc12}{RGB}{128,128,240}
\definecolor{dc13}{RGB}{180,105,255}
\definecolor{dc14}{RGB}{255,0,255}
\definecolor{dc15}{RGB}{212,0,148}
\definecolor{dc16}{RGB}{214,112,218}
\definecolor{dc17}{RGB}{147,20,255}
\definecolor{dc18}{RGB}{70,100,255}
\definecolor{dc19}{RGB}{0,140,255}
\definecolor{dc20}{RGB}{140,230,240}

\def\BibTeX{{\rm B\kern-.05em{\sc i\kern-.025em b}\kern-.08em
    T\kern-.1667em\lower.7ex\hbox{E}\kern-.125emX}}
\begin{document}

\title{High-Precision Fabric Defect Detection via Adaptive Shape Convolutions and Large Kernel Spatial Modeling\\ 
\thanks{*Corresponding author. 

This work was supported by the Shandong Youth University of Political Science Doctoral Research Startup Fund (XXPY23036).}
}

\author{\IEEEauthorblockN{1\textsuperscript{st} Shuai Wang}
\IEEEauthorblockA{\textit{School of Information Engineering} \\
\textit{Shandong Youth University of Political Science}\\
Jinan, China \\
email:230013@sdyu.edu.cn}
\and
\IEEEauthorblockN{2\textsuperscript{nd} Yang Xu}
\IEEEauthorblockA{\textit{School of Information Engineering} \\
\textit{Shandong Youth University of Political Science}\\
Jinan, China \\
email:sdyu2022xy@163.com}
\and
\IEEEauthorblockN{3\textsuperscript{rd} Hui Zheng}
\IEEEauthorblockA{\textit{School of Information Engineering} \\
\textit{Shandong Youth University of Political Science}\\
Jinan, China \\
email:sdyuzh80780443@163.com}
\and
\IEEEauthorblockN{4\textsuperscript{th} Baotian Li*}
\IEEEauthorblockA{\textit{School of Information Engineering} \\
\textit{Shandong Youth University of Political Science}\\
Jinan, China \\
email:sdyu\_lbt@163.com}
}

\maketitle

\begin{abstract}
Detecting fabric defects in the textile industry remains a challenging task due to the diverse and complex nature of defect patterns. Traditional methods often suffer from slow inference speeds, limited accuracy, and inadequate recognition rates, particularly in scenarios involving intricate or subtle defects.
To overcome these limitations, we introduce Fab-ASLKS, an advanced fabric defect detection framework built upon the YOLOv8s architecture. Fab-ASLKS incorporates two key modules: (1) the Adaptive Shape Convolution Module (ASCM), which leverages adaptive shape convolution within the Neck to enhance feature fusion and improve efficiency by extending the capabilities of the standard C2f structure, and (2) the Large Kernel Shift Convolution Module (LKSCM), designed to emulate large kernel effects within the Backbone, enabling superior spatial information extraction. These modules collaboratively optimize feature extraction and information integration across the network.
Extensive experiments conducted on the Tianchi fabric defect detection dataset demonstrate that Fab-ASLKS achieves a 5\% improvement in mAP@50 over the baseline, showcasing its capability to deliver high precision and efficiency. 
\end{abstract}

\begin{figure}[t]
      \centering
     \includegraphics[width=1\linewidth]{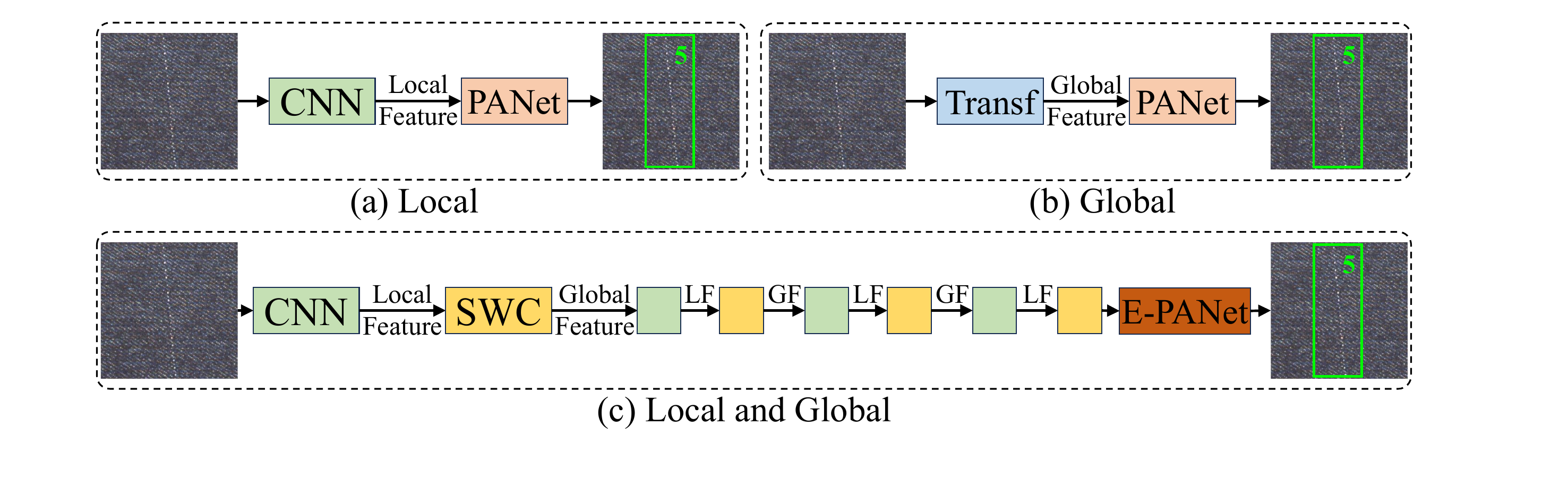}
     \vspace{-.5cm}
    \caption{Different structures employed in fabric defect detection. (a) Traditional convolution for extracting local features. (b) Models based on transformers or large kernel convolutions extract global features. (c) Fab-ASLKS. We employ the original Conv-BN-SiLU (CBS) module in the YOLOv8 model to extract local features and large kernel shift convolution (LKSC) to capture global features. Meanwhile, the Adaptive Shape Convolution Module (ASCM) enhances the fourth C2f module in the Neck (PANet), enabling the dynamic adjustment of convolutional kernels and expanding the receptive field. These enhancements significantly bolster the model's capability to accurately detect intricate defect patterns. \textbf{CNN:} Convolutional Neural Networks, \textbf{PANet:} Path Aggregation Network, \textbf{Transf:} Transformer, \textbf{E-PANet:} Enhanced PANet. }
    \vspace{-.4cm}
    \label{fig:innovations}
\end{figure}

\section{Introduction}

The rapid progress of society has elevated fabric to a cornerstone of daily life and industrial production, with applications spanning clothing, interior design, healthcare, aerospace, and the military~\cite{guoIntelligentQualityControl2024, rasheed2020fabric}. However, fabric defects remain a significant challenge in the textile industry, stemming from machine malfunctions, yarn breakage, or improper handling. Detecting these defects is critical to ensuring product quality. While traditional manual visual inspections are still widely practiced, they are labor-intensive, error-prone, and inefficient, particularly in real-time, high-throughput scenarios. These limitations necessitate the development of automated and efficient defect detection systems capable of handling complex defect patterns.

Initial research efforts gravitated toward two-stage object detection algorithms~\cite{weng2024enhancing, li2024lr, qiao2022novel, weng2023novel}, renowned for their precision and robustness. Models such as Faster R-CNN~\cite{lin2024fabric4show, WOS:000818880500001,WOS:000850266800011, WOS:000647639000012,wang2021patterned, renFasterRCNNRealtime2016} leverage a region proposal network (RPN) to generate candidate regions, followed by refined classification and localization. The stage-wise processing inherent to these methods enables precise identification and localization of defects, significantly reducing error rates compared to manual inspection. However, their limited detection efficiency and suboptimal real-time performance hinder their scalability for industrial applications~\cite{luAnchorFreeDefectDetector2023}.
Recently, diffusion models have gained traction in this domain due to their remarkable capacity for generating high-quality data representations. These models~\cite{shen2024imagdressing, shen2024imagpose} have been effectively applied in areas such as anomaly detection and defect localization, demonstrating robust capabilities in capturing fine-grained patterns across large datasets. By modeling defects as progressive noise reduction tasks, diffusion-based approaches offer a promising alternative to traditional methods, balancing precision with computational efficiency.

To address these challenges, one-stage object detection algorithms, as shown in Fig. \ref{fig:innovations} (a), have gained significant attention due to their superior speed and real-time performance. Models such as YOLO~\cite{wang2024yolo, WOS:001080648000001, WOS:000744582800010, WOS:001021028800010, jocher2020ultralytics} and SSD~\cite{WOS:000884026500001, zhangImprovedMobileNetV2SSDLiteAutomatic2022} eliminate the need for explicit region proposals, directly performing classification and localization in a single stage. This design paradigm facilitates faster processing, making these models suitable for real-time defect detection on modern textile production lines. However, their performance often lags in terms of accuracy and recognition rates, especially for complex defect patterns.
Recently, transformer-based models~\cite{yan2024enhancing,tang2024fourier,zhangMixedAttentionBased2023,liu2024feature} have shown promising results in fabric defect detection by capturing long-range dependencies and contextual information, as shown in Fig. \ref{fig:innovations} (b). Despite their advantages, these models often exhibit high parameter counts and computational complexity, posing significant challenges for real-time deployment in industrial scenarios.

In this paper, we propose {Fab-ASLKS}, a novel fabric defect detection model that addresses the limitations of existing methods and achieves high accuracy in identifying complex textile defects. Fab-ASLKS is built upon the YOLOv8s architecture~\cite{mei2024research} and incorporates two innovative modules to enhance detection performance, as shown in Fig. \ref{fig:innovations} (c). 
The first module, {Adaptive Shape Convolution Module (ASCM)}, is integrated into the Neck to enhance the C2f structure. ASCM introduces adaptive shape convolution (ASC), which dynamically adjusts convolutional kernels to expand the receptive field. This mechanism significantly improves the model's ability to capture spatial transformations, leading to enhanced feature extraction efficiency and accuracy.
The second module, {Large Kernel Shift Convolution Module (LKSCM)}, is incorporated into the Backbone and leverages large kernel shift convolution (LKSC)~\cite{li2024shift} to efficiently simulate the effects of large convolutional kernels. This approach enables the model to capture global context and spatial information more effectively without a substantial increase in computational complexity, which is critical for detecting complex defects.
These enhancements contribute to a 5\% improvement in the mAP@50 metric over the baseline, demonstrating that Fab-ASLKS achieves superior accuracy and efficiency for real-time fabric defect detection in industrial applications.
The primary contributions of this work are as follows:
\begin{itemize}
    \item We introduce the Adaptive Shape Convolution Module (ASCM), which enhances the C2f module in the Neck by incorporating ASC, enabling dynamic adjustment of convolutional kernels and expanding the receptive field. This improves feature extraction efficiency and spatial transformation capture.
    \item We propose the Large Kernel Shift Convolution Module (LKSCM), integrated into the Backbone, which efficiently simulates large convolutional kernels without increasing computational complexity, enhancing the model's ability to capture global context and spatial information.
    \item Experimental results on the Tianchi fabric defect detection dataset demonstrate that Fab-ASLKS achieves a 5\% improvement in mAP@50 compared to the baseline, validating its effectiveness in real-world applications.
\end{itemize}

\section{Related Work}\label{sec:rw}

\subsection{Two-stage Object Detection Algorithms}

Two-stage object detection algorithms first generate candidate regions and then perform fine-grained classification and localization on these regions. Representative methods include the region-based convolutional neural network (R-CNN) series, such as Faster R-CNN~\cite{lin2024fabric4show, WOS:000818880500001,WOS:000850266800011,WOS:000647639000012,wang2021patterned}.  
Cascade Faster R-CNN~\cite{WOS:000818880500001} improves fabric defect detection by using transfer learning with ResNet50 and region of interest (ROI) alignment, integrating a multi-scale feature pyramid, and employing cascaded modules to enhance semantic information and sample distinction.  
SFN R-CNN~\cite{WOS:000850266800011} combines a feature pyramid network (FPN) with ResNet-101 for improved feature extraction and utilizes soft non-maximum suppression (NMS) and ROI alignment to refine region proposals while preserving low-level location information.  
E R-CNN~\cite{WOS:000647639000012} integrates an FPN, deformable convolutional network (DCN), and Distance intersection over union (IoU) Loss into Faster R-CNN, improving detection accuracy and speed.  
Fabric4show~\cite{lin2024fabric4show} enhances fabric defect detection by optimizing region proposals and CNN features, using novel methods like reordering fabric datasets.  
DBPF-RCNN~\cite{wang2021patterned} focuses on single-category defect detection, optimizing region proposals and CNN features to increase flexibility and practicality for real-world scenarios.
Although these algorithms excel in precision, their slower training speeds and higher complexity make them less suitable for real-time industrial applications.

\subsection{One-stage Object Detection Algorithms}

\begin{figure*}[t]
      \centering
     \includegraphics[width=0.95\linewidth]{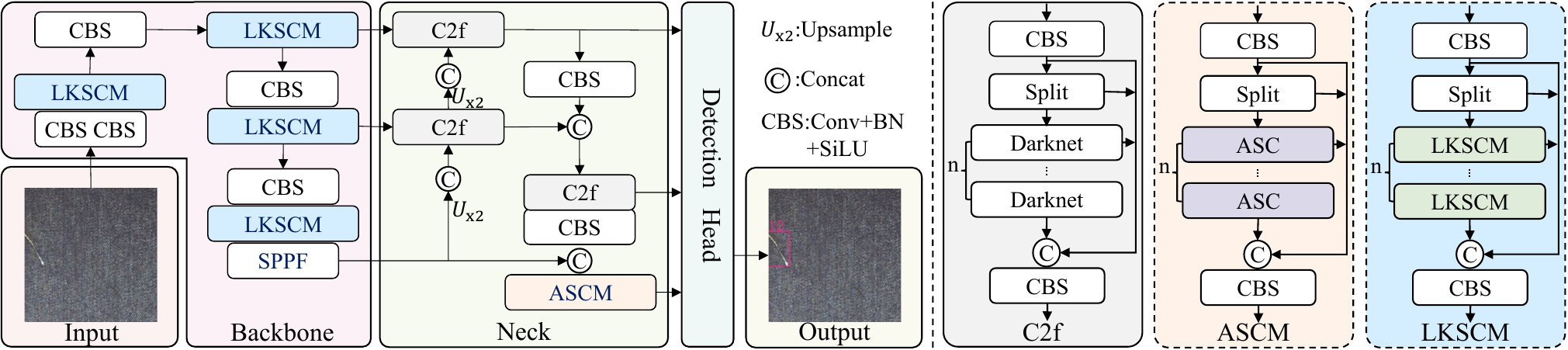}
    \caption{The proposed Fab-ASLKS model. Enhance the C2f module in the Backbone with the large kernel shift convolution (LKSC). Utilize the adaptive shape convolution (ASC) to expand the receptive field of the fourth C2f module in the Neck.}
    \vspace{-.3cm}
    \label{fig:yolov8s}
\end{figure*}

One-stage object detection algorithms perform object localization and classification in a single stage, making them ideal for high real-time performance requirements. Prominent examples include the YOLO series and SSD, which output category probabilities and location coordinates through forward propagation.  
EGNet~\cite{sudha2021robust} mitigates overfitting via data augmentation and max-pooling, enhances gradients using momentum stochastic gradient descent, and employs continuous $1\times1$ convolutional layers for dimensionality reduction and feature fusion, thereby improving efficiency and detection speed.  
YOLOX-CATD~\cite{wang2023fabric} incorporates an improved FPN and pyramid attention network (PAN) in the tiny defect detection layer (TDDL), enhancing localization accuracy for tiny defects.  
AFAM~\cite{wang2023adaptively} introduces an adaptively fused attention module to spatial and channel-wise enhance feature maps, improving attention flow and enriching background information for better detection of small or subtle defects.  
YOLOv8n-LAW~\cite{mei2024research} incorporates large-scale kernel network (LSKNet) attention mechanisms, replaces PAN-FPN with Adaptive FPN, and substitutes complete IoU (CIoU) loss with weighted IoU version 3 (WIoU v3) loss, enhancing feature extraction, semantic alignment, and defect distinction.
Despite innovations in data augmentation, attention mechanisms, feature pyramids, and large-kernel convolutions, one-stage algorithms still face challenges in maintaining high accuracy and recognition rates in complex defect scenarios.

\subsection{Transformer-Based Approaches}

Transformer-based models have recently gained popularity in fabric defect detection due to their ability to capture long-range dependencies and contextual information.  
YOLOvT~\cite{huYOLOvTCSPNetbasedAttention2024} integrates the swin transformer to enhance the receptive field and improve the model's ability to capture global information.  
YOLOv8-ST~\cite{chenFabricDefectDetection2024} embeds a swin transformer block into the C2f module to improve accuracy in detecting small defects. It also uses a bidirectional feature pyramid network (BiFPN) for adaptive feature weighting, a convolutional block attention module (CBAM) to emphasize key features, and the Wise-IoU loss function to address sample imbalance, enhancing detection accuracy and reducing leakage rates.  
MADCAE~\cite{zhangMixedAttentionBased2023} leverages large-kernel convolutions and a hybrid attention module to extract multi-scale features and enrich spatial and channel information, overcoming the limitations of traditional encoders.
While these models significantly improve accuracy and feature representation, their increased complexity and higher parameter counts make them less feasible for resource-constrained industrial environments.

\section{Proposed Method}\label{sec:method}
\subsection{Overview}
 
YOLOv8 consists of four main components: Input, Backbone, Neck, and Head. The input layer scales image data and performs data augmentation for model training. The Backbone extracts features using convolutional layers, residual connections, and bottleneck structures to create feature maps with varying receptive fields. The Neck facilitates multi-scale feature fusion by integrating feature maps from different Backbone stages, enhancing feature representation. Finally, the head performs target detection and classification, predicting objects of various sizes through multi-scale prediction.

The proposed Fab-ASLKS model is an advanced fabric defect detection framework based on the YOLOv8s, as shown in Fig. \ref{fig:yolov8s}, designed to accurately identify complex textile defects. It integrates two key modules: ASCM  and LKSCM. The ASCM enhances the traditional C2f module in the Neck, using ASC to dynamically adjust convolutional kernels, enhancing feature extraction efficiency and accuracy. 
Meanwhile, the LKSCM is incorporated into the Backbone, using LKSC to simulate the effects of large convolutional kernels without increasing computational complexity, thereby improving the model's ability to capture global context and spatial information.

\subsection{Adaptive Shape Convolution Module (ASCM)} 

\begin{figure}[t]
  \centering
  \includegraphics[width=0.9\linewidth]{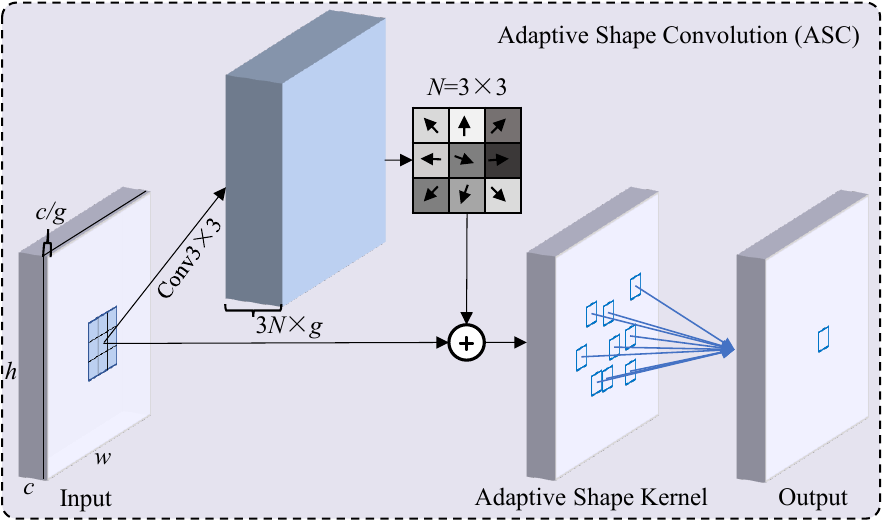}
  \caption{The Framework of adaptive shape convolution (ASC). First, divide the \(c\) channels into \(g\) groups, then generate offsets (\(\Delta m, \Delta n\)) and weights (\(w\)) through \(g\) group convolutions. Next, the offsets are added to the original convolution pixel positions, the convolution is performed, and the weights are multiplied at each position to obtain the ASC output.
  }
  \vspace{-.2cm}
  \label{fig:ASC}
\end{figure}
  
ASC is an efficient operator for various visual applications, as shown in Fig. \ref{fig:ASC}.

\begin{equation}\label{eqn-1}
y = \text{concat}(y_1, y_2, \ldots, y_G).
\end{equation}

Where each group output \( y_g \) is computed as:

\begin{equation}\label{eqn-2}
y_g(p_0) = \sum_{k=1}^{K} w_{g,k} \cdot x_g(p_0 + p_k + \Delta p_{g,k}).
\end{equation}

In this formulation:
\( y \) is the final output feature map after concatenating the outputs from all groups.
\( y_g(p_0) \) is the output feature for group \( g \) at position \( p_0 \).
\( G \) is the total number of groups.
\( K \) is the number of sampling points in each group of the convolutional kernel.
\( w_{g,k} \) is the weight of the \( k \)-th kernel point in group \( g \).
\( x_g \) is the input feature map for group \( g \).
\( p_k \) is the regular grid sampling location.
\( \Delta p_{g,k} \) is the learned offset for the \( k \)-th sampling point in group \( g \).
\(\text{concat}(\cdot)\) denotes the concatenation operation across all groups.
This approach enables ASC to efficiently capture complex spatial transformations using grouped convolutions and flexible sampling locations. The ASC module comprises an ASC convolution layer, a batch normalization layer, and the SiLU activation function. 

The ASCM is implemented by replacing the darknet module in the C2f section of the Neck with the ASC module. 
Initially, the input feature map \( X \) is processed through a convolutional layer, resulting in \(\text{Conv}(X)\), which is then divided into two segments, \( X_1 \) and \( X_2 \).

\begin{equation}\label{eqn-3}
X_1, X_2 = \text{Split}(\text{Conv}(X)).
\end{equation}

The segment \( X_2 \) is processed through an ASC module, producing the feature map \( Y_2 \).

\begin{equation}\label{eqn-4}
Y_2 = \text{ASC}(X_2).
\end{equation}

Subsequently, \( Y_2 \) is further refined by passing it through \( n-1 \) additional ASC modules, resulting in \( Y_2' \).

\begin{equation}\label{eqn-5}
Y_2' = \underbrace{\text{ASC}(\text{ASC}(\ldots \text{ASC}(Y_2) \ldots))}_{n-1 \text{ times}}.
\end{equation}

The features \( X_1 \), \(\text{Conv}(X)\), \( Y_2 \), and \( Y_2' \) are concatenated along the channel dimension. This concatenated feature set is then passed through a final convolutional layer to generate the output feature map.

\begin{equation}\label{eqn-6}
\text{Output} = \text{Conv}(\text{Concat}(X_1, \text{Conv}(X), Y_2, Y_2')).
\end{equation}

Introducing ASC blocks into the C2f module significantly enhances feature fusion efficiency. This design enables the model to effectively capture complex spatial transformations.

\subsection{Large Kernel Shift Convolution Module (LKSCM)} 

\begin{figure}[t]
    \centering
   \includegraphics[width=0.98\linewidth]{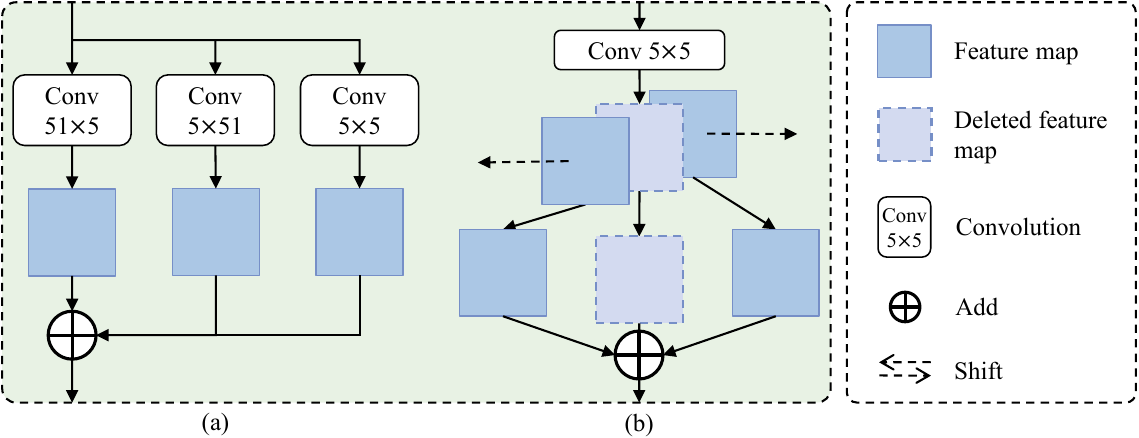}
   \vspace{-.2cm}
  \caption{The Framework of large kernel shift convolution (LKSC). (a) Decompose the $51\times51$ convolution into three convolutions: $51\times5$, $5\times51$, and $5\times5$, using SLaK. (b) Decompose the $51\times5$ convolution into eleven $5\times5$ convolutions using shift convolution (SC), followed by fusion after shifting.}
  \vspace{-.3cm}
  \label{fig:lksc}
\end{figure}
  
LKSC is an efficient convolutional neural network(CNN) architecture that uses a lightweight shift operation to partially replace traditional convolution computations. This approach achieves spatial information exchange through simple feature map shifts, eliminating the need for additional parameters or multiplications. The primary advantage of LKSC is its ability to significantly reduce computational complexity and parameter count while maintaining model performance. This architecture excels in resource-constrained environments, making it suitable for applications requiring efficient inference. By integrating pointwise convolutions and nonlinear activation functions, LKSC accelerates inference without compromising accuracy.
The formula for the LKSC module is as follows:
\begin{multline}\label{eqn-7}
    y(p_{(i,j)}) = \prod_{k=0}^{f(kw,kh,A) } \sum_{m=0}^{A} \sum_{n=0}^{A} 
    w(p_{(\Delta m,\Delta n)}+\Delta p) \cdot \\
    x(p_{(i,j)}+p_{(\Delta m,\Delta n)}  +\Delta p),
\end{multline}

where:$y(p_{(i,j)})$ denotes the output feature map at position $(i,j)$.
$\prod$ signifies all predetermined non-overlapping small convolutions.
$A$ represents the size of the small convolution kernel.
$kw$ and $kh$ denote the focus length and focus width, respectively.
$f(kw,kh,A)$ is a function related to $(kw,kh,A)$.
$\Delta m = m - \frac{A}{2}$ and $\Delta n = n - \frac{A}{2}$ represent the offsets from the center of the small convolution kernel.
$\Delta p=g(kh,k)$ is a function related to the weight and feature offset $(kh,k)$.

The LKSCM is implemented by replacing the darknet module in the C2f section of the Backbone with the LKSC. This enhancement strategy retains the original architecture's advantages while incorporating performance improvements from the shift operator. Integrating the LKSC into the C2f module significantly reduces convolution operation complexity and time cost.

\section{Experiment and Analysis}\label{sec:exp} 
To demonstrate the effectiveness of the proposed Fab-ASLKS framework, we compare it against several state-of-the-art fabric defect detection methods using the large-scale Tianchi fabric dataset.

\subsection{Datasets}

\begin{table}[t]
  \centering
  \caption{Distribution of defects in the dataset.}\label{table:samples}
  \begin{tabular}{@{}ccccc@{}}
  \toprule
  No. & Color & Name &  Train Set Image &  Val Set Image       \\ \midrule
  1  &\cellcolor{dc1}& holes                 & 409   & 93   \\
  2  &\cellcolor{dc2}& water stains, etc.    & 609   & 158  \\
  3  &\cellcolor{dc3}& three-yarn defects    & 1550  & 267  \\
  4  &\cellcolor{dc4}& knots                 & 2561  & 465  \\
  5  &\cellcolor{dc5}& pattern skips         & 640   & 153  \\
  6  &\cellcolor{dc6}& hundred-leg defects   & 804   & 130  \\
  7  &\cellcolor{dc7}& neps                  & 282   & 57   \\
  8  &\cellcolor{dc8}& thick ends            & 452   & 85   \\
  9  &\cellcolor{dc9}& loose ends            & 827   & 152  \\
  10 &\cellcolor{dc10}& broken ends          & 590   & 124  \\
  11 &\cellcolor{dc11}& sagging ends         & 333   & 50   \\
  12 &\cellcolor{dc12}& thick fibers         & 1293  & 219  \\
  13 &\cellcolor{dc13}& weft shrinkage       & 861   & 150  \\
  14 &\cellcolor{dc14}& sizing spots         & 1398  & 329  \\
  15 &\cellcolor{dc15}& warp knots           & 451   & 82   \\
  16 &\cellcolor{dc16}& star skips, etc.     & 621   & 129  \\
  17 &\cellcolor{dc17}& broken spandex       & 589   & 115  \\
  18 &\cellcolor{dc18}& color shading, etc.  & 1972  & 390  \\
  19 &\cellcolor{dc19}& abrasion marks, etc. & 1210  & 220  \\
  20 &\cellcolor{dc20}& dead folds, etc.     & 789   & 169  \\\bottomrule
  \end{tabular}
  \vspace{-0.30cm}
  \end{table}

The Tianchi fabric dataset contains 5,913 defective fabric images, each sized \(2446 \times 1000\) pixels. It includes 20 types of defects, as shown in Table \ref{table:samples}: holes; stains (water, oil, dirt); three-yarn defects; knots; pattern skips; hundred-leg defects; neps; thick ends; loose ends; broken ends; sagging ends; thick fibers; weft shrinkage; sizing spots; warp knots; star skips and skipped wefts; broken spandex; unevenness, waves, and color shading; abrasion, rolling, repair marks; and dead folds, cloud weaves, double wefts. 

\begin{figure}[t]
  \centering
 \includegraphics[width=0.9\linewidth]{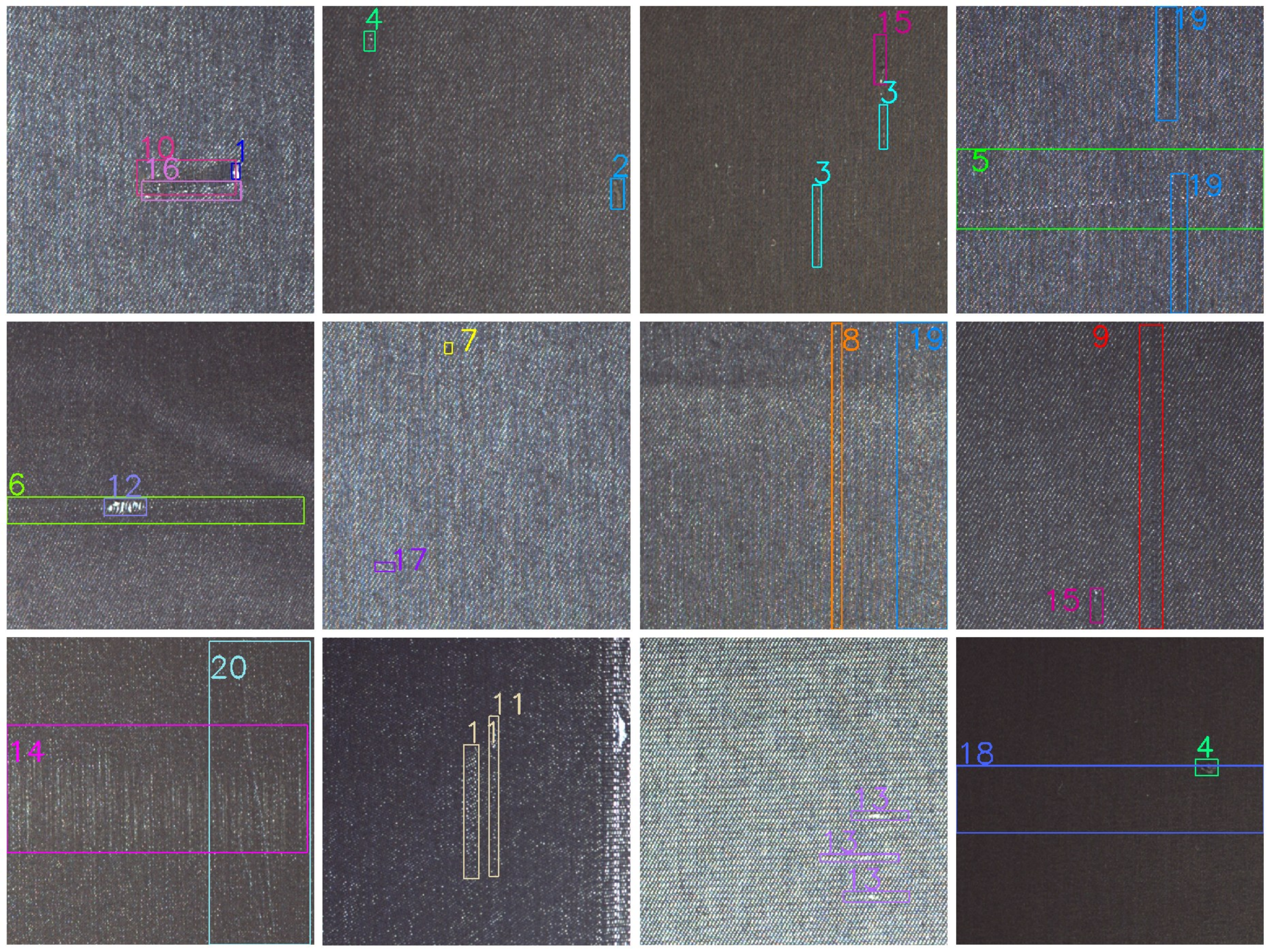}
\caption{Example images of 20 types of defects with ground truth bounding boxes}
\vspace{-.4cm}
\label{fig:sample}
\end{figure}

To enhance detection performance, each defect image is cropped into eight \(640 \times 640\) pixel samples, resulting in a total of 7,170 images. The experiment uses data augmentation techniques like random flipping, scaling, rotation, gaussian blur, and color jitter. The dataset is divided into a training set and a validation set in a 4:1 ratio. The training set consists of 5,736 images, and the test set contains 1,434 images. Fig. \ref{fig:sample} showcases examples of the 20 types of defects. Inspection reveals that the dataset contains many large and long defects, significantly increasing detection complexity.

\subsection{Evaluation Metrics} 

The mean average precision at IoU threshold 0.50 (mAP@50) is a widely used evaluation metric in object detection tasks. It measures a model's prediction precision by considering the IoU between predicted and ground truth bounding boxes. Calculate the average of the AP values across all classes at an IoU threshold of 0.50. This is expressed as:

\begin{equation}\label{eqn-9}
\text{mAP@50} = \frac{1}{N} \sum_{i=1}^{N} \text{AP}_i,
\end{equation}

where \( N \) is the number of classes and \( \text{AP}_i \) is the average precision for class \( i \).

The mAP@50 provides a comprehensive measure of a model's detection performance, balancing precision and recall across all classes at a specific IoU threshold, thus offering insights into both the accuracy and robustness of the model's predictions.

\subsection{Implementation Details}
Experiments were conducted on a server equipped with eight RTX 4090 GPUs and AMD EPYC 7T83 CPUs. Each training instance utilized a 16-core CPU, 32GB of memory, and an RTX 4090 GPU. The YOLOv8 model was tailored for fabric defect detection, processing input images of \(640 \times 640\) pixels. Training was executed over 500 epochs with a batch size of 16, employing the SGD optimizer with a learning rate of 0.005 and a momentum of 0.937. To mitigate overfitting, a weight decay of 0.0005 was applied. A 3-epoch warmup phase was implemented to stabilize the initial training process. Mosaic data augmentation was used to enhance detection across various scales and orientations, ensuring robust model performance. Training was halted if performance indicators did not improve after 50 epochs, with the average training duration being approximately 200 epochs.

\subsection{Comparison with State-of-the-art Methods}

\begin{table}[t]
  \centering
  \caption{Comparison with state-of-the-art methods on Tianchi fabric dataset. A dash ("-") denotes unreported results in the respective paper.}\label{table:tb2}
  \begin{tabular}{{@{}ccccc@{}}}
      \toprule
  Methods            & mAP@0.5 (\%) & Param. (M)   & Year \\
   \midrule
  Faster R-CNN~\cite{renFasterRCNNRealtime2016}       & 35.90 &   25.60    & 2016 \\
  Tood~\cite{feng2021tood}                            & 44.10 &   53.26    & 2021 \\
  YOLOv8n-LAW~\cite{mei2024research}                         & 49.80 &   -        & 2024 \\
  YOLOv5s~\cite{jocher2020ultralytics}                & 50.20 &   14.40    & 2020 \\
  CA-FPN~\cite{luAnchorFreeDefectDetector2023}          & 53.10 &   36.81    & 2023 \\
  YOLOX-CATD~\cite{wang2023fabric}                    & 54.63 &   27.15    & 2023 \\
  AFAM~\cite{wang2023adaptively}                      & 56.70 &   69.63    & 2023 \\
  YOLO-TTD~\cite{yue2022research}                     & 56.74 &   24.52    & 2022 \\
  \hline
  Baseline                                            & 57.40 &   11.10    & 2023 \\
  Fab-ASLKS                                              & \textbf{60.30}     & \textbf{10.50}   &   Ours 
  \\ \bottomrule
\end{tabular}
\vspace{-0.20cm}
\end{table}

Table \ref{table:tb2} presents a comprehensive comparison of our Fab-ASLKS model against state-of-the-art fabric defect detection algorithms, evaluated on the Tianchi fabric dataset. Performance is primarily measured using mAP@50.

Our Fab-ASLKS model achieves an mAP@50 of 60.3\%, outperforming all other methods listed in the table. This result highlights the superior accuracy and robustness of our approach in detecting fabric defects. Notably, the Fab-ASLKS model surpasses the baseline with a 5\% improvement in mAP@50. This improvement underscores the effectiveness of integrating ASC and LKSC into the YOLOv8s framework, enhancing feature extraction and spatial information capture.

Compared to YOLOX-CATD and AFAM, which achieve mAP@50 values of 54.63\% and 56.70\% respectively, our model offers higher accuracy and maintains a lower parameter count of 10.50 million. This balance of performance and efficiency makes Fab-ASLKS suitable for real-time applications in resource-constrained environments.

Overall, the results in Table \ref{table:tb2} validate our enhancements, positioning Fab-ASLKS as a leading solution for fabric defect detection in the textile industry.

\subsection{Ablation Studies and Analysis}  

\begin{figure}[t]
  \centering
 \includegraphics[width=0.9\linewidth]{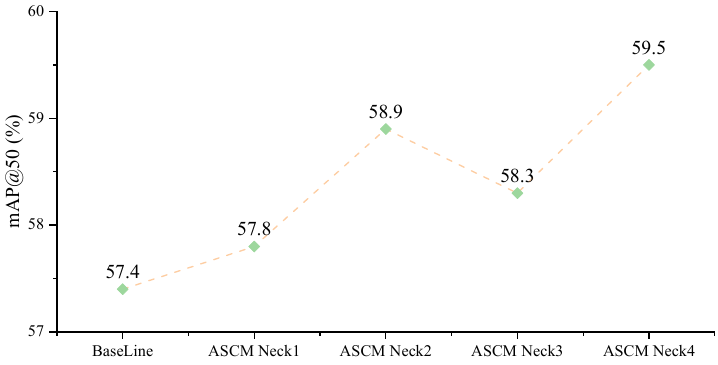}
\caption{Comparison of effects by replacing different C2f modules in the Neck with ASCM. Here, ASCM Neck$i$ denotes the improvement of the $i$-th C2f module from top to bottom in the Neck using ASCM.}
\label{fig:ASCdiffpos}
\end{figure}

\begin{figure}[t]
  \centering
 \includegraphics[width=0.9\linewidth]{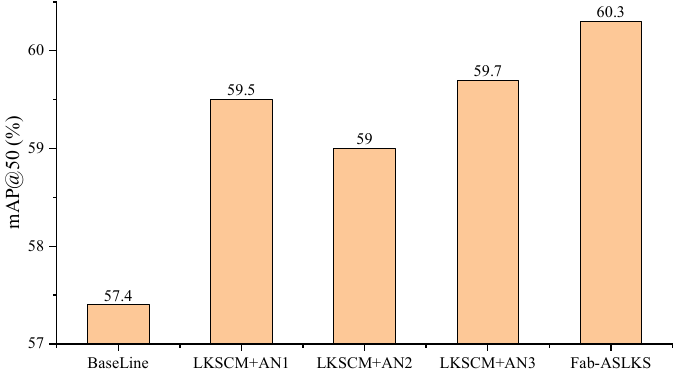}
\caption{Comparison of model performance with simultaneous addition of LKSCM and ASCM at different positions. AN$i$ denotes the improvement of the $i$-th C2f module from top to bottom in the Neck using ASCM.}
\label{fig:lkscaddASCdiffpos}
\end{figure}

\begin{table}[t]
  \centering
  \caption{The ablation experiments for proposed modules based on the principle of control variables, including ASCM and LKSCM. The "$\times$" indicates that the module was not added in this set of experiments.}\label{table:tb3}
  \begin{tabular}{{@{}ccccc@{}}}
    \toprule
  \bf Baseline  &\bf ASCM Neck4  &\bf LKSCM &\bf mAP@50 &\bf mAP@50-95 \\
  \midrule
   \checkmark & \texttimes & \texttimes &57.4 &29.4  \\
   \checkmark & \checkmark & \texttimes &59.5 &31.0  \\
   \checkmark & \texttimes & \checkmark &59.4 &31.2  \\
   \checkmark & \checkmark & \checkmark &60.3 &31.8  \\
  \bottomrule
\end{tabular}
\end{table}

\begin{figure}[t]
  \centering
 \includegraphics[width=0.8\linewidth]{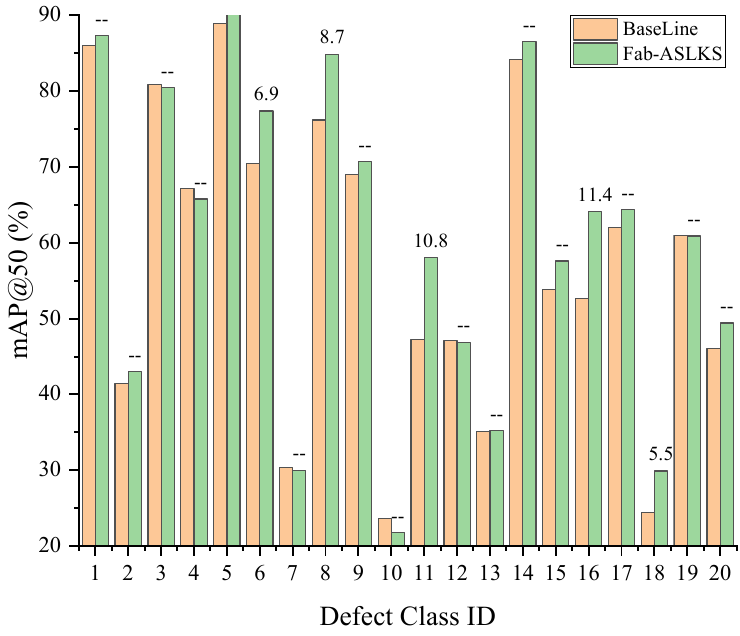}
\caption{Comparison of Detection Results for Each Defect Type Between Baseline and Fab-ASLKS Models. The figure also highlights defect types where the Fab-ASLKS model exceeds the baseline by more than a 5\% difference in the mAP@50 metric.}
\label{fig:eachClassmAP50}
\end{figure}

The comparison results in Fig. \ref{fig:ASCdiffpos}, Fig. \ref{fig:lkscaddASCdiffpos}, Fig. \ref{fig:eachClassmAP50}, and Table \ref{table:tb3} demonstrate that the proposed Fab-ASLKS method is superior to many state-of-the-art fabric defect detection methods.
The proposed Fab-ASLKS method is comprehensively analyzed from four aspects to investigate the logic behind its superiority.
\subsubsection{\textbf{Role of ASCM} }
Fig. \ref{fig:ASCdiffpos} shows that strategically integrating the ASCM module can significantly boost model performance, especially when optimally positioned within the network architecture. This highlights the importance of module placement in achieving optimal detection outcomes.
\subsubsection{\textbf{Influence of LKSCM}}
Fig. \ref{fig:lkscaddASCdiffpos} shows that strategically integrating the ASCM, particularly in the fourth position within the Neck, significantly boosts model performance. Combining with LKSCM further amplifies these gains, underscoring the importance of module placement and synergy in achieving optimal detection outcomes.
\subsubsection{\textbf{Impact on Model Accuracy}}
Table \ref{table:tb3} presents ablation experiment results evaluating the impact of the ASCM and LKSCM modules on the Fab-ASLKS model's performance. The baseline model achieves an mAP@50 of 57.4\% and an mAP@50-95 of 29.4\%, serving as a reference point. Introducing the ASCM in the fourth position of the Neck increases the mAP@50 to 59.5\% and mAP@50-95 to 31.0\%, demonstrating its effectiveness in enhancing feature extraction and detection accuracy. Similarly, the LKSCM module enhances performance, achieving an mAP@50 of 59.4\% and an mAP@50-95 of 31.2\%, indicating its effectiveness in reducing computational complexity while preserving spatial information capture. Integrating both modules yields the highest performance, with an mAP@50 of 60.3\% and an mAP@50-95 of 31.8\%, highlighting their complementary nature. These results underscore the significant contributions of ASCM and LKSCM to the enhanced performance of the Fab-ASLKS model, achieving state-of-the-art results in fabric defect detection.
\subsubsection{\textbf{Large and Long Defects Performance Analysis}}

\begin{figure*}[t]
  \centering
 \includegraphics[width=0.95\linewidth]{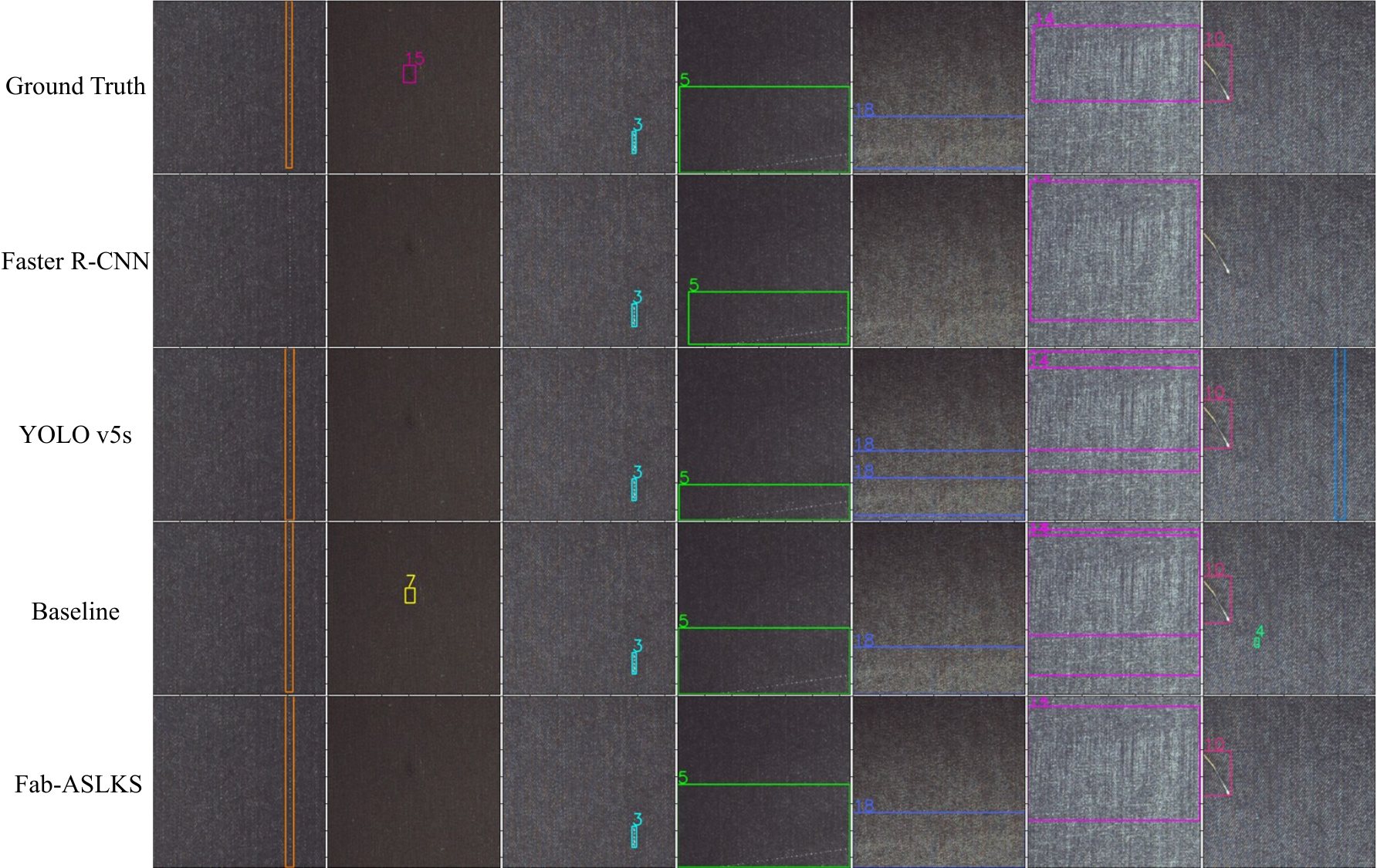}
\caption{Comparison of algorithm detection results. Compared our method with Faster R-CNN, YOLOv5s, and baseline. The numbers near the boxes in the figure represent the category numbers of the defects, and the category names and colors can be found in the Name and Color column of Table~\ref{table:samples}.}
\label{fig:compare}
\end{figure*}

Fig. \ref{fig:eachClassmAP50} illustrates the comparative detection performance between the Baseline and Fab-ASLKS models across various defect types, revealing an overall improvement in the mAP@50 metric for most categories. Notably, the most significant gains are observed in large and long defects, such as "holes," "pattern skips," and "broken ends," as depicted in Fig. \ref{fig:sample}. These improvements underscore the Fab-ASLKS model's enhanced capability to capture complex spatial features and transformations, attributed to the integration of the ASCM and LKSCM modules. These modules bolster the model's feature extraction efficiency, enabling it to better handle the intricate details and spatial relationships inherent in such defects. The consistent performance gains across diverse defect types highlight the model's robustness and adaptability.

\subsection{Visualization}  

Figure \ref{fig:compare} illustrates the detection results of our Fab-ASLKS model compared to other state-of-the-art models, highlighting its superior performance. 
The figure clearly shows that our algorithm achieves better detection results, with fewer false positives and false negatives, and detection boxes more closely aligned with the ground truth. 
For instance, in the 5th and 6th sets of detection results, other algorithms exhibit issues with duplicate detections. 
In the 7th set, our algorithm correctly identifies the 10th type of defect, whereas the Baseline and YOLOv5s models encounter false detection issues, and the Faster R-CNN algorithm misses the detection entirely. The Fab-ASLKS model consistently achieves higher detection accuracy across various defect types, excelling in handling complex scenarios with intricate textures and overlapping defects. This is due to the integration of the ASCM and LKSCM, which enhance spatial feature extraction. 
The model shows marked improvements in detecting challenging defects, such as and elongated ones, which are often inaccurately detected by other models. 
This capability is crucial for applications requiring high precision and reliability. Furthermore, the Fab-ASLKS model maintains consistent performance across a wide range of defect types, underscoring its versatility and adaptability. Unlike other models that may excel in specific categories, our model delivers high accuracy and robustness across the board, establishing it as a leading solution for fabric defect detection.

\section{Conclusion}\label{sec:con} 
In this work, we introduced Fab-ASLKS, a novel model designed to address the challenges of fabric defect detection. Built upon the YOLOv8s architecture, Fab-ASLKS integrates two key innovations: the Adaptive Shape Convolution Module (ASCM) and the Large Kernel Shift Convolution Module (LKSCM). ASCM enhances the C2f module in the Neck by dynamically adjusting convolutional kernel shapes, thereby expanding the receptive field and improving feature extraction efficiency. LKSCM, integrated into the Backbone, simulates large kernel effects to capture spatial information more effectively. Our extensive experiments on the Tianchi fabric defect detection dataset demonstrate a 5\% improvement in mAP@50 over the baseline, underscoring the model's superior accuracy and efficiency. These results establish Fab-ASLKS as a robust solution for real-time fabric defect detection in industrial applications. Future work will focus on further optimizing the model's architecture and exploring its application to other types of industrial defect detection tasks.

{\small 
\bibliographystyle{IEEEtran}
\bibliography{ref}
}
\end{document}